\documentclass{article}

\usepackage{times}
\usepackage{graphicx} 
\usepackage{caption}
\usepackage{subcaption}

\usepackage{natbib}

\usepackage{algorithm}
\usepackage{algorithmic}

\usepackage{amsmath}

\usepackage{booktabs}
\usepackage{multicol}
\usepackage{multirow}
\usepackage[flushleft]{threeparttable}
\usepackage{todonotes}
\usepackage{hyperref}

\newcommand\abs[1]{\left|#1\right|}

\usepackage[group-separator={,}]{siunitx}
\usepackage[accepted]{whi2016}

\icmltitlerunning{Interactive Semantic Featuring for Text Classification}

\begin{document} 

\twocolumn[
\icmltitle{Interactive Semantic Featuring for Text Classification}
\setcounter{page}{51}

\icmlauthor{Camille Jandot$^1$, Patrice Simard$^1$}{camille.jandot@telecom-paristech.fr~~patrice@microsoft.com}
\icmlauthor{Max Chickering$^1$, David Grangier$^2$, Jina Suh$^1$}{\{dmax,jinsuh\}@microsoft.com~~grangier@fb.com}
\icmladdress{$^1$ Microsoft Research. Redmond, WA, USA.~~~~~$^2$ Facebook AI Research.  Menlo Park, CA, USA}

\icmlkeywords{interactive learning,interactive featuring,dictionary smoothing}

\vskip 0.3in
]

\begin{abstract} 
In text classification, dictionaries can be used to define 
human-comprehensible features.  We propose an improvement to
dictionary features called {\em smoothed dictionary features}. These features
recognize document {\em contexts} instead of n-grams. We
describe a principled methodology to solicit dictionary features from
a teacher, and present results showing that models built using these
human-comprehensible features are competitive with models trained with
Bag of Words features.
\end{abstract} 

\section{Introduction}

Featuring for machine learning (ML) is a way for the model builder, or
the \textit{teacher}, to describe the data and encode pertinent
signals to the underlying learner so that the machine can ``see'' the
data in the way the teacher intended.  In the case of text
classification, the ``Bag of Words'' (BoW) representation--where each
feature represents a single word or n-gram in the document--is a
popular choice; these features are usually generated automatically
from text corpora and require no input from the teacher. Although BoW
features have shown to be very successful in experiments and in
practice \cite{croft2010search,scott1999feature}, the dimensionality
of the space (i.e., the number of words used) is so large that the
resulting model is hard to understand and diagnose for
humans. Furthermore, the BoW features do not take the order of words
into account, which means that it necessarily loses information about
its local context, semantic meaning, or syntactic
structure \cite{scott1999feature} (e.g. ``social security in media''
and ``security in social media'' are indistinguishable by BoW).

As an alternative to BoW features, {\em dictionaries} (or {\em
 lexicons}) allow the teacher to provide further supervision.
 Dictionaries contain multiple n-grams that convey the same meaning or
 are semantically related to each other. Dictionaries can be generated
 manually by the domain expert \cite{taboada2011lexicon} or by
 leveraging external sources such as
 WordNet \cite{fellbaum1998wordnet}. By using manually-created
 dictionaries, we can improve generalization \cite{williams2015fast}
 and interpretability of the resulting model; furthermore, because the
 dimensionality of the feature space can be much lower, we require far
 fewer examples to train the models. Dictionary features also have a
 number of drawbacks: they can be time-consuming and tedious for
 teachers to construct, they can result in low-recall models if the
 teacher omits an important concept, and they have the same
 lack-of-ordering-information limitation as discussed above for BoW
 features.

In this paper, we propose an improvement to the dictionary
feature, called the {\em smoothed} dictionary features, that
takes into account the context in which a word appears. Instead of
recognizing the {\em explicit occurrences} of n-grams of the
dictionaries in the document, the smoothed dictionary features
leverage the {\em probability} that a word belongs to the dictionary
given its context (i.e. the words that co-appear with the considered
word). We describe a procedure that interleaves featuring and labeling and show that models built using the resulting features
compare favorably to models built using BoW features for two
classification tasks.

\section{Semantic Featuring}

In this section, we describe two ways for a teacher to communicate 
semantic meaning to the model, 
dictionary features and smoothed dictionary
features, and an interactive teaching process that can be used to
solicit these features.

\subsection{Dictionary Features}

A {\em dictionary} $D$ is a set of n-gram terms $\{T_1, \ldots,
T_k\}$. For a given text document, we say that the dictionary {\em
  matches at a position} if the document contains some $T_i \in D$
at that position. A {\em dictionary feature} for a document is an
aggregate count of the dictionary matches within that document. We
assume for the remainder of the paper that, for dictionary $D_i$
with $N_i$ matches in document $d$, the feature value $F_i$ is defined as
$$
F_i(d) = log(1 + N_i)
$$
Note that the feature value is not a function of the identity of the
matching n-grams. This means that care must be taken when constructing
the dictionary. 

On the positive side, this allows faster generalization by
considering the presence of all n-grams to be equivalent. On the
negative side, this equivalence multiplies the risk that a dictionary
would fire outside of its intended scope. In particular, a single
n-gram can derail the semantic intent of the dictionary.  For example,
a dictionary containing the months of the year will include the term
``May'', which will match on ``May I help you''.

Dictionaries, sometimes referred to as {\em lexicons} or
{\em gazetteers}, have shown
promise for both sentiment analysis \cite{taboada2011lexicon},
 for extraction tasks such as named-entity recognition \cite{smith2006using}, 
 and for utterance analysis tasks \cite{williams2015fast}.

\subsection{Smoothed Dictionary Features} \label{smoothed}

We propose an improvement to the dictionary feature that, instead of
using {\em literal} dictionary matches, predicts the {\em probability}
that there is a dictionary match according to a context model. By
using the contexts instead of dictionary matches, our model can better
capture the semantics instead of the syntax of the dictionary terms;
the corresponding feature will match n-grams that are not in the
dictionary but have similar contexts, and the feature will be
suppressed for n-grams that are in the dictionary but have unrelated
contexts.

Our context model is a logistic-regression model built to output, for
any given n-gram $g$ within a document, the probability that $g$
belongs to the dictionary $D$. We train the model by selecting
positive and negative instances of the dictionary terms, in context, from a
large, unlabeled corpus of text. Each context feature $c$ of the context model
corresponds to the sequence of unigrams within a {\em context window}
near $g$, with the requirement that these unigrams do not overlap with
$g$. For example, one context window might consist of the five tokens
following $g$. Given the $i^{th}$ context window, let $U^i = \{u_1, ...,
u_k\}$ denote the set of unigrams in that window for some document and
position of $g$; we define the value of the corresponding
context feature $c^i$ via the log odds
\begin{equation} \label{eq:feature}
c^i = log \frac{p(g \in D | u_1, ..., u_k)}{p(g \not \in D | u_1, ..., u_k)}
\end{equation}
We use a na\"{\i}ve-Bayes classifier for the probabilities in Equation
\ref{eq:feature}. In our experiments, we used ten context features
corresponding to non-overlapping windows of increasing size around
$g$: a size-one window immediately before and after $g$, a size-two
window immediately before and after the two respective size-one
windows, a size-four window before and after the size-two windows, and
so on.

The logistic-regression model combines the context-window features
from Equation \ref{eq:feature} together via:
$$
p(g \in D | Context) = \frac{1}{1 + e^{-\theta_0 + \sum_{i} c^i \cdot \theta_i}}
$$ By combining the na\"{\i}ve-Bayes classifier scores together with
logistic regression, our model benefits from the efficiency of
na\"{\i}ve-Bayes for both training and runtime--requiring only
pair-wise counts to produce the scores--without needing to rely its
independence assumptions in the final prediction.

Given the model $p(g \in D_i | Context)$ and a threshold $\gamma_i$, we
say that the dictionary $D_i$ {\em smooth matches at a position} if,
for the n-gram $g$ at that position we have $p(g \in D_i | Context)
\geq \gamma_i$.  A {\em smoothed dictionary feature} for a document is an
aggregate count of the smooth dictionary matches. As above we assume
for the remainder of the paper that, for a dictionary $D_i$ with $N_i$
{\em smooth} matches in document $d$, the {\em smoothed} feature $M_i$
is defined as
$$
M_i(d) = log(1 + N_i)
$$ 
The threshold $\gamma_i$ can be set manually by the teacher by
looking at sorted lists of contexts as we discuss below. In our
experiments, however, we used an unsupervised method of setting
$\gamma_i$ such that we had a constant average number of smoothed
matches per document across a large corpus of unlabeled documents.

In Table \ref{table:context}, we show how, unlike dictionary features,
the smoothed dictionary features can distinguish homonyms. The table
shows a list of contexts from the the Open Directory Project (ODP)
data set (http://www.dmoz.org) where the unigram
``may'' occurs, ranked by decreasing probability that ``may'' occurs
in the dictionary $D_{month}$ (containing all 12 months, including
``may'') according to the context model trained as described above. A teacher 
would likely to manually choose a value between $0.0024$ and $0.0052$ as 
the threshold $\gamma_{month}$.

\begin{table*}
\caption{``May'' in context for $D_{months}$}
\vskip 0.15in
\begin{center}
\begin{footnotesize}

\begin{tabular}{@{}c@{}c@{}r@{}c@{}l@{}}

  \toprule
  
  \textsc{ Perc. } & \textsc{ Prob. } & \textsc{ Words before} & \textsc{ Targ. } & \textsc{Words after }  \\ \midrule
  
  $0.0$ & $1.0000$ & \footnotesize{ september 2008 august 2008 july 2008 june 2008} & \footnotesize{ may} & \footnotesize{  2008 april 2008 march 2008 february 2008 january} \\
  $2.5$ & $0.0200$ & \footnotesize{ as bachelor of theology when , on 24} & \footnotesize{ may} & , \footnotesize{ , 1323 , john xxii at the request} \\
  $5.0$ & $0.0052$ & \footnotesize{ were arrested . the following year , in} & \footnotesize{ may} & \footnotesize{ , four thousand huguenots assembled on the seine}\\
  $7.5$ & $0.0024$ & \footnotesize{  0 like 6 5 blog gear about rss} & \footnotesize{ may} & \footnotesize{ 13 macro madness usa - washington posted by} \\
  $10.0$ & $0.0018$ & \footnotesize{ to the new syndicomm web site as you} & \footnotesize{ may} & , \footnotesize{ or may not have noticed already , i }  \\ 
  $12.5$ & $ 0.0017$ & \footnotesize{ being phased out . the mp command set} & \footnotesize{ may} & \footnotesize{ or may not ( according to the implementation}  \\
  $15.0$ & $0.0013$ & , \footnotesize{ not be married to one another , who} & \footnotesize{ may} & \footnotesize{ or may not both be the biological parent} \\   
  $17.5$ & $0.0012$ & \footnotesize{ many cases to ` art ' films that} & \footnotesize{ may}  & \footnotesize{ appease only the film  connaisseur ' and} \\ 
  $20.0$ & $0.0012$ & \footnotesize{  based on the filing system date , which } & \footnotesize{ may} & \footnotesize{ or may not be accurate ) arm object} \\
  $22.5$ & $0.0012$ & \footnotesize{  convenience , in the same spirit as you} & \footnotesize{ may} & \footnotesize{ read a journal or a proceedings article in }   \\ 
 \bottomrule

\end{tabular}

\begin{tablenotes}
    \item{\scriptsize{Note: The table above displays the occurrences of the target term ``may'' and the context in which it appears, sorted by probability that the considered occurrence is part of the dictionary. If the threshold of the dictionary is set to $0.0024$, the percentage of words that will trigger the dictionary is $7.5\%$.}}
\end{tablenotes}
\end{footnotesize}
\end{center}
\label{table:context}
\end{table*}

The context model can also be used to recommend n-grams to {\em add}
to existing dictionaries. In particular, for any n-gram $g \not \in D$,
we can average the context-model predictions $p(g \in D | Context)$
over all instances of $g$; we can then suggest to add the
highest-scoring n-grams to $D$. To illustrate this, we
started with an incomplete dictionary of months $D_{month'}$ =
[``january"; ``february"; ``march"] and averaged the
context-model predictions for all unigrams; the top-scoring unigrams
were ``april", ``august", ``october", ``september", ``july",
``november", ``june", and ``december'', followed by some misspelled
months, abbreviations of months, ``may'' and months in German and French. 

\subsection{Soliciting Features From the Teacher} \label{generatefeatures}

When training models with BoW features, the teacher needs only to
supply labels (likely through an active-learning strategy
\cite{settles2009active}) to train the model, as the feature space is
pre-defined. If our goal is to build classifiers with semantic
features such as dictionary or smoothed dictionary features,
we need a methodology to solicit them from the teacher.
Although the teacher could pre-define all the semantic features 
before concentrating on labeling of documents, we take instead
the approach of interleaving featuring and labeling. This allows the
teacher to re-examine feature decisions during the labeling effort,
which can prevent the teacher from making incorrect assumptions about
the data distribution and classification errors. 

\begin{algorithm}[htb]
\begin{small}
\caption{Interactive Teaching Process}
\label{alg:interactive}
\begin{algorithmic}[1]
\REPEAT 
  \IF{(Not feature blind)}
    \STATE {Sample, label, and add to training}
  \ELSE
    \STATE{Add or modify feature}
  \ENDIF
  \STATE{Train the model}
\UNTIL{$\abs{learner - teacher} < \epsilon$}
\end{algorithmic}
\end{small}
\end{algorithm}

Consider the teaching process shown in Algorithm~\ref{alg:interactive}. At each iteration, the teacher samples for an example to add to the training set. Each iteration will train a new model, and if the resulting model is confused by the new example due to missing features, the model has a "feature-blindness". The teacher then adds a feature to help distinguish the confusing patterns until the blindness is resolved. 

A good teaching strategy is one that uses this teaching process with two key principles. (1) \textit{Sample for confusion}. During the Sample step, a good teacher checks for generalization ability by searching for examples that are likely to confuse the student. This active sampling strategy helps reduce the number of labels. (2) \textit{Feature for generalization}. During the Feature step, a good teacher chooses features that will not only resolve existing conflicts but also resolve future conflicts, thus improving generalization on unseen data. For instance, a feature blindness for ``piano" could benefit from adding an entire concept class for ``musical instruments". Features chosen in this fashion are not likely to lead to overtraining because they are only added when necessary. The resulting features are interpretable because there are relatively small in number (i.e. limited to teacher's interaction) and they are expressed directly by the teacher. 


\section{Experiments}
\label{experiments}

In this section we compare classifiers built using (1) BoW features,
(2) dictionary features, and (3) smoothed dictionary features. We use the
documents (web pages) and labels provided by ODP data set,
 which we split randomly into two sets
$\mathcal{S}$ and $\mathcal{S}_{test}$, containing respectively
\num[group-separator={,}]{330398} and \num[group-separator={,}]{140839}
 documents. The first set is used to build
the dictionaries and to train the models, and the second set is used
for testing. We chose two classification tasks,  \textit{health} and \textit{music}, that are similar in their positive distribution in $\mathcal{S}_{test}$ (\num[group-separator={,}]{5987} and \num[group-separator={,}]{5988} documents are positive, respectively).

\subsection{Dictionary Generation}
We built dictionaries as
described in Section~\ref{generatefeatures} using an interactive
learning interface \cite{simard2014ice} that enables us to both label
and add (not smoothed) dictionary features. We first created an
initial dictionary that contains descriptive words about the main
classification task (e.g., for {\em music}, we used the dictionary:
[``music''; ``musician''; ``musical'']). We then started labeling
items by searching the unlabeled corpus for certain keywords 
or by soliciting suggested items to label. 
Suggested items were selected using active learning strategies designed to identify 
documents with potential feature deficiencies, including uncertainty 
sampling \cite{settles2009active} and disagreement sampling where 
documents were selected based on the disagreement between the current model and a model
trained with BoW features. Throughout the process, the model was 
 trained in the background every time a few labels were submitted.

When an error occurs in the training data (i.e., the
classifier disagrees with our label for an example that is in the
training set), we tried to identify a feature blindness and addressed the 
error either by refining an existing dictionary or adding a new one. We also
refined existing dictionaries using the context models as described in
Section~\ref{smoothed}; for example, the first suggested words for
the dictionary [``piano''; ``guitar''; ``sax''; ``bass''] were
``bassoon", ``saxophone", ``timpani" (a synonym of kettledrums),
``cello", ``tuba", ``oboe", ``clarinet", ``ibanez" (a well-known
guitar brand), ``viola" and ``trombone".

The process produced $27$ dictionaries for \textit{health} within $2.4$ hours (we estimate that $44\%$ of time was spent adding features and $56\%$ labeling), and $24$ dictionaries for \textit{music} within $1.7$ hour ($31\%$ of which was spent adding features).

\subsection{Model Training}
For each of the three candidate feature sets and classification tasks,
we trained a model using L2-regularized logistic regression. For the 
BoW features, we used TF-IDF weighting. In order to eliminate the 
potential effect of different featuring strategies in our training data, 
we retained only the features from the process described above and 
created a separate training data set for evaluation.

We initialized the
{\em common} training set with $5$ positive and $5$ negative examples
drawn from $\mathcal{S}$ and then applied uncertainty sampling using
the BoW classifier to iteratively choose a document whose predicted score 
is closest to $0.5$ to build the training set $\mathcal{S}_{train}$, totaling 300 documents with 50/50 class balance. Note that because this training set was constructed
explicitly to improve the BoW model, the semantic-feature models are
somewhat at a disadvantage. 

\begin{figure}
  \center
  \includegraphics[width=1\columnwidth]{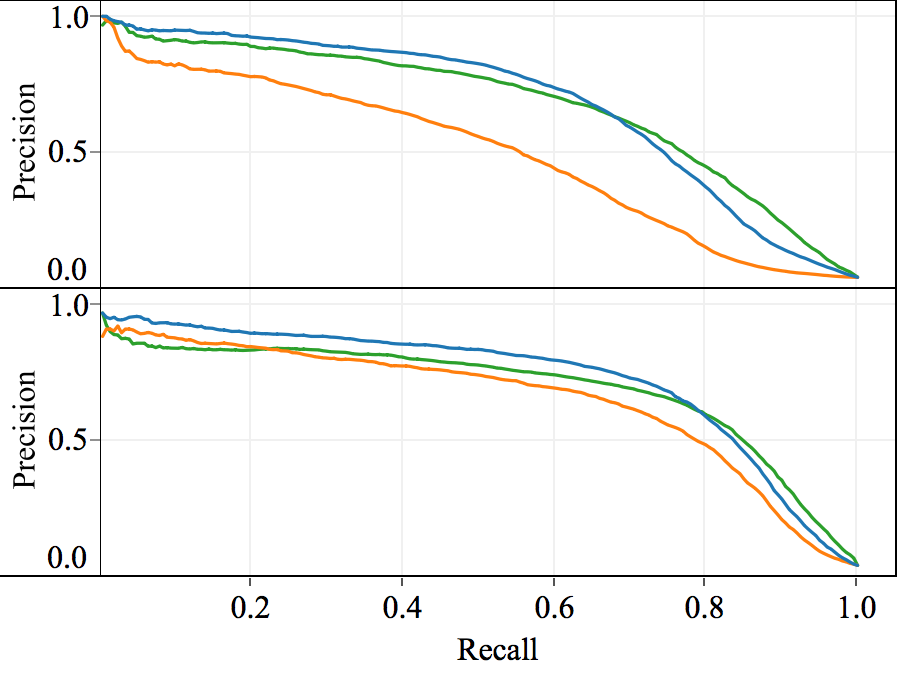}
  \caption{Precision-Recall curves using BoW (blue), dictionary (orange), and smoothed dictionary (green) features for \textit{health} (top) and \textit{music} (bottom) classifiers.}
\label{fig:precisionrecall}
\end{figure}

\begin{table}[H]
\centering
\caption{AUROC and accuracy of the classifiers}
\vskip 0.15in
\begin{center}
\begin{small}
\begin{sc}
\begin{tabular}{rcccc}
  \toprule
\multicolumn{1}{l}{}         & \multicolumn{2}{c}{Health}               & \multicolumn{2}{c}{Music} \\ \midrule
\multicolumn{1}{r|}{Feature} & AUC($\%$) & \multicolumn{1}{c|}{ACC($\%$)} & AUC($\%$)    & ACC($\%$)    \\ \midrule
\multicolumn{1}{r|}{BoW}    & 93.3   & \multicolumn{1}{c|}{97.5}   & 95.3     & 97.7 \\
\multicolumn{1}{r|}{Dict}     & 86.4   & \multicolumn{1}{c|}{96.5}   & 94.1     & 97.2 \\
\multicolumn{1}{r|}{Smoo}  & 95.1    & \multicolumn{1}{c|}{97.3}   & 96.7    & 97.4  \\ \bottomrule
\end{tabular}
\end{sc}
\end{small}
\end{center}
\label{table:res}
\end{table}

\subsection{Results}
Figure~\ref{fig:precisionrecall} and Table~\ref{table:res} illustrate that the model built using smoothed dictionary features performs competitively with BoW model. 
However, the compressed BoW (trained with L1 with negligible loss of performance) model uses 6932 (resp. 6999) for health (resp. music), which become intractable for humans to understand, while only 24 (resp. 27) weights are used for dictionaries. When we force BoW model to use the same number of words as dictionaries (189 and 185 words), we see that BoW performs much worse, as expected. The top 100 BoW words (sorted by weight magnitude) are easy to interpret. After this, the effect of removal or addition of words is very difficult to predict. Dictionaries are semantically linked to “feature blindness”. The effect of removal or addition is more predictable because it is linked to being the last feature that allows the model to distinguish two subsets of examples of different classes.

\section{Discussion}

In this paper, we introduced smoothed dictionary features as well as
an interactive teaching process. While our preliminary experimental
results shows that smoothed dictionaries perform competitively with BoW
features, much work remains to be done.  The use of smoothed
dictionary features is one way to incorporate contextual information
into the model. We need to perform a series of experiments
comparing smoothed dictionary feature to other context models in terms
of overall model performance as well as interpretability and
cohesiveness of the features.  We introduced Algorithm 1 as a process
for gathering semantic features. It remains to be seen if constructing
features in an interactive loop is indeed more efficient in terms of
labeling effort. We also need to explore whether or not the features
constructed in this process are more interpretable.  In a followup
experiment, instead of using a fixed $\gamma_i$, we would like to solicit
the teacher for the appropriate threshold.

\section*{Acknowledgments}
CJ is affiliated with T\'{e}l\'{e}com ParisTech. Part of this work was done when DG was at Microsoft Research.

\bibliography{reference}
\bibliographystyle{icml2016}

\end{document}